\documentclass[letterpaper, 10pt, conference]{ieeeconf}
\IEEEoverridecommandlockouts
\usepackage[T1]{fontenc}
\usepackage{amsmath,amssymb}
\usepackage{graphicx}
\usepackage{booktabs}
\usepackage{array}
\usepackage{float}
\usepackage{cuted}
\usepackage{placeins}
\usepackage{url}
\usepackage{xcolor}
\usepackage{tikz}
\usetikzlibrary{arrows.meta,positioning,calc,shapes.geometric,backgrounds,fit,decorations.pathreplacing}
\usepackage{pgfplots}
\pgfplotsset{compat=1.18}
\definecolor{kwarm}{HTML}{C8571E}\definecolor{kwarmsoft}{HTML}{F7E2D4}
\definecolor{kcool}{HTML}{1E5BB8}\definecolor{kcoolsoft}{HTML}{DCE6F6}
\definecolor{kmute}{HTML}{5C6F63}\definecolor{kmutesoft}{HTML}{E1E7E2}
\definecolor{kgrey}{HTML}{8B939C}\definecolor{kpaper}{HTML}{F4F5F1}

\newcommand{\KPITitle}{KPI: A Promptable Kernel for Physical Interaction on Humanoids}
\newcommand{\KPIAuthors}{%
  Yikai Wang$^{1}$, Honghao Zhu$^{2}$, Xiao Hu$^{3}$, Hao Zhang$^{1}$,\\
  Zelin Wang$^{2}$, Yip Fun Yeung$^{2}$, Ding Zhao$^{1}$, and Lingfeng Sun$^{2}$}
\newcommand{\KPIAuthorNotes}{%
  \thanks{$^{1}$Carnegie Mellon University. $^{2}$Autel US. $^{3}$Northeastern University.}%
  \thanks{\raggedright Corresponding authors: Yikai Wang (\mbox{yikaiw2@andrew.cmu.edu}) and Lingfeng Sun (\mbox{lingfengsun1996@gmail.com}).}%
}

\title{\LARGE\bfseries \KPITitle}

\author{\KPIAuthors\KPIAuthorNotes}

\IEEEaftertitletext{\vspace{-3.5mm}\centerline{\includegraphics[width=\textwidth]{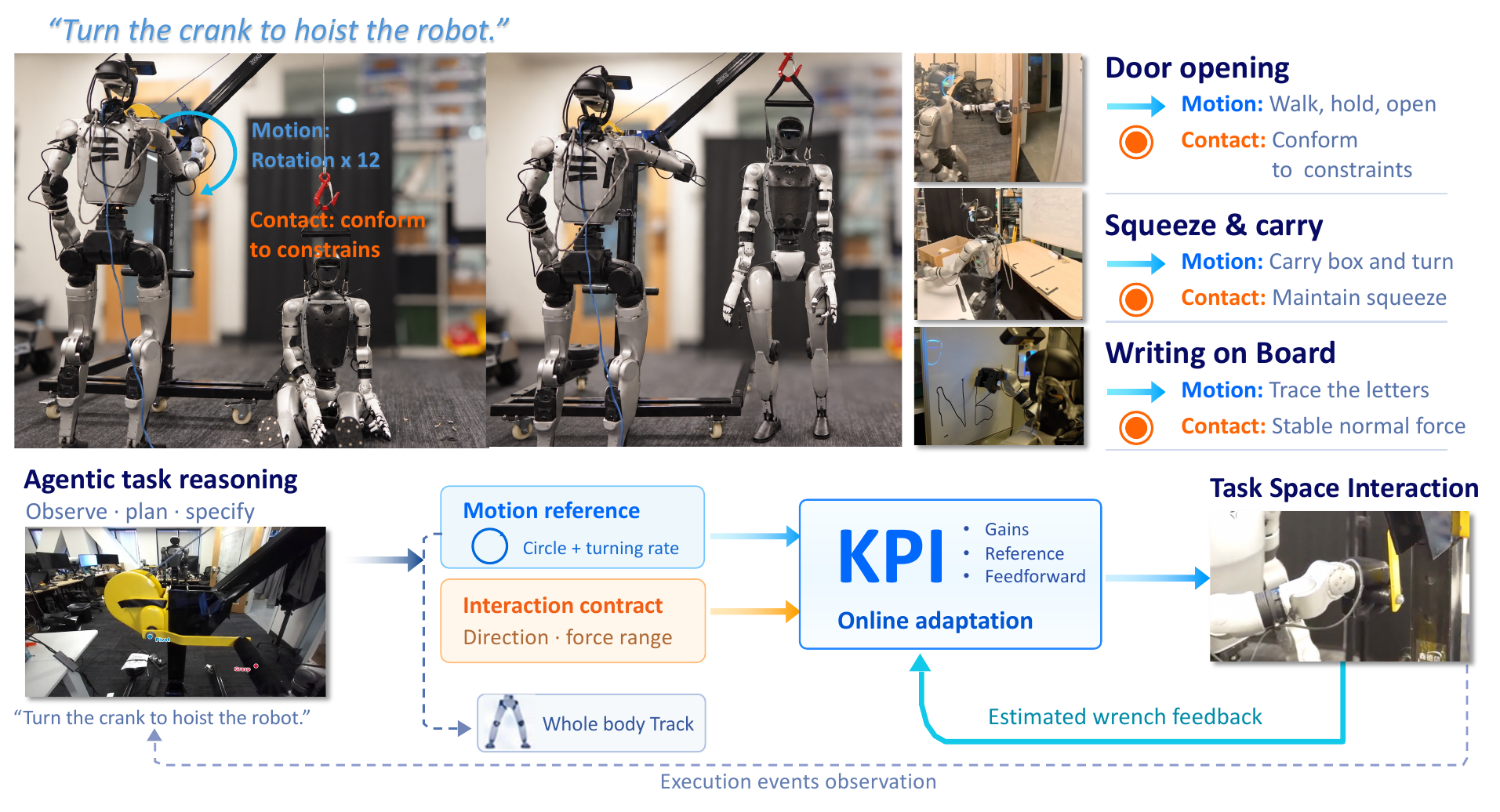}}\vspace{0.5mm}
  \parbox{\textwidth}{\footnotesize\refstepcounter{figure}\label{fig:overview}Fig.~\thefigure.\quad \textbf{KPI:} a promptable kernel for physical interaction. KPI uses motion references, interaction contracts, and estimated wrench feedback to adapt the arm controller, while an unmodified whole-body tracker controls the legs and waist. From one instruction, a humanoid turns a hand-winch to hoist a second robot. Further agentic demonstrations include door opening and box carrying.}\vspace{3mm}}

\begin{document}
\bstctlcite{KPI:BSTcontrol}

\maketitle
\thispagestyle{empty}
\pagestyle{empty}

\begin{abstract}
Humanoids now walk, balance and reach with remarkable generality: one whole-body tracking policy follows references from a human, or from an end-to-end policy. That generality travels in the trajectory, and a trajectory alone carries limited information about the interaction it should produce: at contact, the executing controller determines how the robot behaves. Single-task policies usually reach hard interactions by optimising trajectory and controller together in simulation; general stacks usually assume a preset or hand-chosen controller. We present KPI, a promptable kernel for physical interaction between the trajectory source and an unmodified whole-body tracker. Instead of a controller fixed before the task, the trajectory source sends a contract: per direction, track, comply, or hold a force range. From tracking error and a wrench estimate, the kernel adapts the arms' stiffness, damping, reference and feedforward toward it at contact rate. We demonstrate KPI through an agentic framework: from one instruction, a vision-language agent writes both the reference trajectory and the contract, with no task-specific code. We demonstrate instruction-driven winch operation, door opening, and box transport, alongside scripted surface-interaction experiments. In the winch demonstration, the humanoid is able to turn a crank to hoist a second robot fully off the ground. Project website: \url{https://kpi-robot.github.io/}.
\end{abstract}

\section{Introduction}

One whole-body tracking policy now carries a humanoid through walking, balancing and reaching, following a reference from a teleoperator~\cite{omnih2o,homie,twist2}, a retargeted human recording~\cite{sonic,gmt,beyondmimic} or a vision-language-action (VLA) model~\cite{gr00tn1,psi0,openhlm}. Consider asking such a system to turn a hand-winch. The commander supplies a circle and a turning rate. It does not know the resistance at each angle of the handle, the exact path the mechanism allows, or how both change as the load rises. The arm meets those conditions with gains that were fixed before the task was known, and millimetres of reference error become tens of newtons against the handle's circle. Today this is handled in one of two ways: a policy trained per task, with the trajectory and the contact response optimised together in simulation~\cite{cola,doorman,hdmi,softa}, or a general stack whose arm gains are chosen before the task is known~\cite{tunetolearn}.

A compliance chosen before the task does not resolve this~\cite{chip,softmimic,lac,mcc,hmc}. On a winch, the arm must be stiff along the handle's tangent and soft along its radius, and that assignment rotates with the handle. Slip, jamming and contact loss develop within tens of milliseconds, below the deliberation time of whoever sets the gains. On a walking humanoid both ends of the contact move, so only quantities measured at the interface keep their meaning. We therefore let the commander state the objective rather than the gain. A force range means the same thing at every deflection, while the force a given stiffness delivers does not. Force ranges are the instance we use here, and other interface quantities can play the same role.

We present KPI, a kernel for physical interaction\footnote{Both senses are intended: the commander states an objective the controller is measured against, and \emph{kernel} is used as in an operating system, for the component that owns the real-time loop and the hardware and exposes them in the terms of the layer above.}, which sits between the commander and the whole-body tracker and leaves that tracker unmodified (Fig.~\ref{fig:overview}). The commander sends a nominal reference together with a \emph{contract}: for each direction, whether to track, comply or hold a force range, and which controller parameters may change. KPI solves for the parameters that meet the contract at 100\,Hz, reading the pose error, the velocity and an estimated wrench. Three parts have to hold together: a contract that a person or a model can state and check, an arm that can carry it out, and the optimization that connects them, which runs during the contact rather than before it. For the arms, we realize Cartesian impedance through the G1's existing per-joint PD interface by decomposing damping into joint-local terms implemented in the motor drivers and cross-joint terms computed on the host.

This gives the commander tolerance. A reference need only be right to centimetres, the geometry can be observed in part, and the wrench can be estimated without a force sensor. The commander names the direction, the force range and which channel may move, and never writes a gain value. The approach also asks little of the layers around it: the whole-body tracker is unmodified, the arm uses the robot's existing joint interface, no wrist force sensor is added and no policy is retrained. A vision-language model can serve as the commander: it grounds the scene from egocentric RGB-D, writes the reference and the contract from an instruction, and revises both after each stage. The interaction is promptable in this sense: it is specified in language rather than in gain values.
We evaluate on a Unitree~G1 with no wrist force sensor, on interactions that differ in kind: a winch, a door, a bimanual box carry, a board written on, and a drawer. A vision-language agent drives the winch, the door and the box from one instruction, with no task-specific code. KPI completes all five trials on each of the three tasks. Across the same tasks, the tracker’s native joint PD arm controller and the sensorless admittance baseline integrated with the same tracker complete zero and one of 15 trials, respectively. The agent's own choices show what the interface buys: asked for stiffness values when lifting a box, it chose 500 and 600\,N/m and drove the arms into motor faults; asked for force ranges on the same task, it chose 25--40\,N and 35--50\,N and lifted the load. On the winch it turned the handle until a second G1 was lifted off the ground. Sec.~\ref{sec:experiments} also compares the kernel with its parameters frozen (KPI-fixed), and reports drawer opening and box carrying while kneeling, standing and running under teleoperation. Our contributions are:
\begin{enumerate}
  \item \textbf{Contract-driven adaptation}. One bounded parameter space and a contract through which the same optimization adapts gains, reference and feedforward during the contact, so that a task is stated as objectives rather than gains.
  \item \textbf{An impedance arm beside an unmodified whole-body tracker}. A Cartesian impedance realized over the robot's per-joint gain interface, with joint-local damping closed in the motor drivers, running while the tracker walks.
  \item \textbf{A vision-language agentic system} that decomposes instructions, grounds the scene, writes references and contracts, and revises them from execution outcomes, executing loco-manipulation without task-specific demonstrations or code.
\end{enumerate}

\begin{figure*}[!t]
  \centering
  \includegraphics[width=\textwidth]{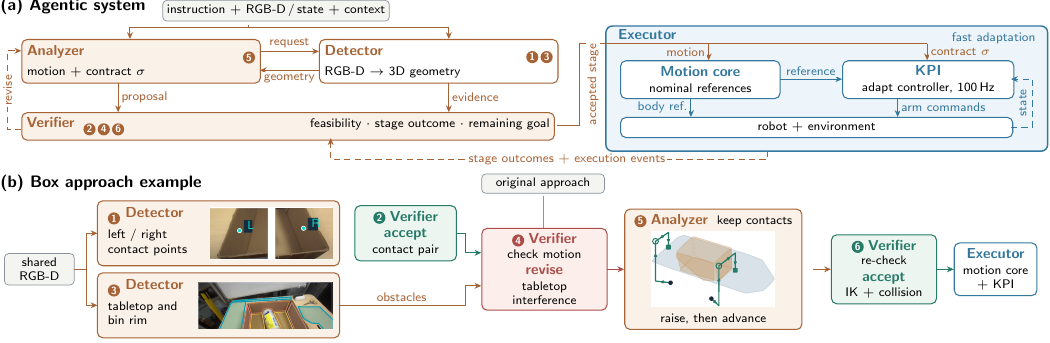}\unskip\par
  \caption{\textbf{Agentic execution through KPI.} (a)~Analyzer writes the motion and the contract $\sigma$ from Detector's geometric evidence, and Verifier checks feasibility, stage outcomes and the remaining goal, returning rejected proposals for revision. Executor runs an accepted stage: a motion core supplies the nominal reference and KPI adapts the arm controller under $\sigma$ from the interaction state. (b)~One revision in the box task, numbered as in (a): Detector locates the two contact faces and the surrounding obstacles, Verifier rejects the original straight approach because it interferes with the tabletop, and Analyzer raises the hands before advancing while keeping both contacts.}
  \label{fig:agent-pipeline}
\end{figure*}

\section{Related Work}\label{sec:related}

\textbf{Trajectory sources and task policies.}
Today's general humanoid stacks are trackers trained on retargeted human motion~\cite{sonic,gmt,beyondmimic} that take their reference from teleoperation~\cite{omnih2o,twist2} or from a humanoid VLA~\cite{psi0,openhlm}. Below the trajectory they share a position-servo arm whose gains are chosen for learnability and fixed at deployment~\cite{tunetolearn,sonic}; HOIST notes that its policy acts ``without explicitly adapting the applied force''~\cite[Sec.~6]{hoist}, and force-aware VLAs so far run on fixed-base arms with force sensing~\cite{forcevla2}. Where the contact response is learned instead, it is learned per task, with trajectory and contact response optimised together in simulation to carry boxes~\cite{cola}, open doors~\cite{doorman,hdmi} or reproduce contact-rich human motion~\cite{contactmimic}, and contact entering training as a reward or a privileged input rather than as a run-time measurement. We build on SONIC~\cite{sonic} unmodified, and a new interaction enters as a contract rather than as a new skill.

\textbf{Task specification.}
Specifying which task directions are governed by motion and which by force is well established: hybrid position/force control specifies a task in a frame attached to the contact~\cite{mason1981,raibertcraig1981}, and constraint-based formulations extend this to geometric constraints under uncertainty~\cite{itasc}. A contract keeps that structure and changes what is stated in it: a range rather than a setpoint, together with the parameters that may move to meet it.

\textbf{Compliance on humanoids.}
The closest work gives a humanoid a compliant response and differs from ours in who sets it, when, and what has to change to change it. Compliance is trained into the tracker as a stiffness or force-threshold command~\cite{softmimic,lac} or as a fixed impedance distilled from a teacher~\cite{ceer}; CHIP, on the same G1 as ours, exposes one compliance coefficient per end-effector, set by the operator or fixed in a VLA's training data~\cite{chip}. MCC estimates the contact wrench from motor current without force sensors and runs a task-space admittance with fixed gain magnitudes on standing robots~\cite{mcc}; HMC blends position, impedance and hybrid-force experts through a behaviour-cloned router~\cite{hmc}. In each, the stiffness is chosen before the contact begins, and SoftMimic names as future work ``how to best select stiffness for a given task''~\cite[Sec.~V]{softmimic}. Where compliance is adapted during execution instead, the adaptation is a residual policy trained per task in simulation, on a fixed-base arm with force sensing~\cite{admittanceresidual}. We take MCC's estimator as our starting point and add the path from the estimated wrench back to the parameters during the contact.

\textbf{Language and agents.}
Agentic humanoid systems ground the scene with a VLM or open-vocabulary segmentation~\cite{sam3}, as we do with a VLM, and send poses down to a whole-body controller or an MPC~\cite{handoff,humanoidcoa}, in the code-as-policy tradition~\cite{cap,voxposer}. Where a language model sets the compliance itself, it writes per-axis stiffness into policy code~\cite{genchip}, sets it about once a second on a fixed arm~\cite{compliantvla}, or retrieves gains from a database on a G1~\cite{humanoidvlm,safehumanoid}; between its decisions the interaction runs on the gains it chose. The kernel takes the same instruction and, instead of gains, a contract, and closes the loop at contact rate.

\section{Method}\label{sec:method}

KPI makes the controller's interaction response an explicit part of the command interface. Alongside a nominal reference, the commander supplies a \emph{contract} stating the interaction objectives. The kernel adapts the controller parameters toward it during execution and returns the interaction state. We describe the contract-driven adaptation, its realization on a humanoid, and its use by a vision-language agent; a teleoperator or a VLA policy uses the same interface.

\subsection{KPI: Contract-Driven Physical Interaction}\label{sec:layer}
\textbf{Controller parameters and online optimization.}
KPI operates on a shared parameter space $\Theta$. In our realization,
\begin{equation}\label{eq:theta}
\theta=(\mathbf{K}_p,\ \mathbf{K}_d,\ \boldsymbol{\rho},\ \mathbf{F}_{\mathrm{ff}})\in\Theta,
\end{equation}
where $\mathbf{K}_p$ and $\mathbf{K}_d$ set stiffness and damping, $\boldsymbol{\rho}$ conditions reference advancement along the nominal trajectory, and $\mathbf{F}_{\mathrm{ff}}$ supplies an additive wrench. For bimanual interaction, $\theta$ collects both arms' parameters. Cartesian impedance realizes this response on our humanoid (Sec.~\ref{sec:controller}).

At each adaptation cycle, KPI uses the interaction state
\begin{equation}\label{eq:interaction-state}
\mathbf{s}_t=(\mathbf{x}^{\star}_t,\ \mathbf{x}_t,\ \mathbf{v}_t,\ \hat{\mathbf{W}}_t),
\end{equation}
where $\mathbf{x}^{\star}_t$ denotes the nominal end-effector reference, including its pose trajectory, desired velocity and geometry; $\mathbf{x}_t$ and $\mathbf{v}_t$ are the current end-effector pose and velocity; and $\hat{\mathbf{W}}_t$ is the estimated external wrench. All quantities use consistent coordinate frames and collect both hands when needed. Under a commander-supplied contract $\sigma$, adaptation is formulated as the constrained update
\begin{equation}\label{eq:adaptation}
\begin{aligned}
\theta_{t+1}\in{}&\operatorname*{arg\,min}_{\theta\in\Theta_\sigma(\mathbf{s}_t)}
\Big[\mathcal{L}_\sigma(\theta;\mathbf{s}_t)
+\lambda\|\theta-\theta_t\|_R^2\Big]\\
\text{s.t.}\quad& f_i^{\min}\le\tilde f_i(\theta;\mathbf{s}_t)\le f_i^{\max},
\quad i=1,\ldots,m.
\end{aligned}
\end{equation}
Each specification contributes an objective term, a force constraint, or both. Here $\mathcal{L}_\sigma$ collects the objectives and $\Theta_\sigma\subseteq\Theta$ restricts changes to the parameters the contract selects, within fixed controller bounds. The immediate force prediction linearizes the impedance response at fixed measured pose and velocity, $\tilde f_i(\theta;\mathbf{s}_t)=\hat f_i+\nabla_\theta f_i^{\top}(\theta-\theta_t)$. A separate damping-dominated, short-horizon surrogate predicts tracking errors (Supplementary Sec.~\ref{sec:supp-gain}); neither requires identifying environmental stiffness. The estimated wrench anchors the force prediction and is not an optimization variable. The bounds apply to the signed robot-on-environment component along a unit direction and to the norm over a subspace. The regularizer holds parameters where the contract leaves them free, with $\lambda>0$ and a positive-definite $R$ accounting for their different units.

\textbf{Interaction contract.}
The commander selects the directions to configure and assigns a specification to each direction or subspace requiring a distinct response:
\begin{equation}\label{eq:contract}
\begin{aligned}
\sigma&=\{\sigma_i\}_{i=1}^{m},\\
\sigma_i&=(d_i,\ o_i,\ [f_i^{\min},f_i^{\max}],\ c_i).
\end{aligned}
\end{equation}
The four terms have the following meanings:
\begin{itemize}
\setlength{\itemsep}{0pt}
\item \emph{Direction $d_i$:} a trajectory tangent or its normal subspace, a plane tangent subspace or its normal, or a unit direction in a declared frame. Directions are resolved from the nominal motion and the current end-effector poses.
\item \emph{Requirement $o_i\in\{\mathrm{tracking},\mathrm{compliant},\mathrm{constrained}\}$:} tracking penalizes motion-tracking errors; compliant favors a target low directional stiffness; constrained imposes the force range without an additional tracking or stiffness-preference objective. A free direction can be assigned either tracking or compliant behavior.
\item \emph{Force range $[f_i^{\min},f_i^{\max}]$:} the lower and upper force bounds for the selected direction or subspace. The interval constrains the response for any requirement; equal bounds specify a target force, and an unbounded interval omits the force restriction.
\item {\raggedright \emph{Channels $c_i$:} a nonempty subset of $\{\mathrm{gain},\mathrm{reference},\mathrm{feedforward}\}$ specifying which parameters may change jointly. These correspond to gain matrices $(\mathbf{K}_p,\mathbf{K}_d)$, reference conditioning $\boldsymbol{\rho}$, and feedforward wrench $\mathbf{F}_{\mathrm{ff}}$, respectively.\par}
\end{itemize}
Here $m$ counts the directions or subspaces configured at one time, and all selected parameters are optimized jointly. In bimanual carrying, for example, the two palm forces have separate bounds while both arms' parameters enter the same problem. Rotation follows the same form, with directions read as axes and force bounds as torque bounds. The directions and channels listed here are those used in this work, not an exhaustive vocabulary.

For a tracking specification, $\tilde{\mathbf e}_i(\theta;\mathbf{s}_t)$ collects two predicted pose errors at $h=10$\,ms: the nominal reference minus the predicted end-effector pose, and the conditioned execution reference minus that pose, resolved along $d_i$ (Supplementary Sec.~\ref{sec:supp-gain}). The objective is
\begin{equation}\label{eq:contract-objective}
\begin{aligned}
\mathcal L_\sigma(\theta;\mathbf{s}_t)
={}&\sum_{i:\,o_i=\mathrm{tracking}}
\|\tilde{\mathbf e}_i(\theta;\mathbf{s}_t)\|_{Q_i}^2\\
&+\sum_{i:\,o_i=\mathrm{compliant}}w_i\|\mathbf K_{p,i}-\mathbf K_{p,i}^{\mathrm{soft}}\|_F^2.
\end{aligned}
\end{equation}
The matrix $\mathbf K_{p,i}$ is the stiffness restricted to the selected direction or subspace, and $\mathbf K_{p,i}^{\mathrm{soft}}$ is its target low stiffness for compliant behavior, a fixed kernel default rather than a contract term. The cost encourages the hand to follow the nominal motion while penalizing reference--hand separation, allowing the execution reference to slow when the hand lags. A constrained specification contributes only its force bounds in~\eqref{eq:adaptation}, so stiffness is retained while the range is satisfied. Parameters not selected by any active specification remain fixed.

The selected channels determine how the optimization can meet these requirements. In our winch experiments, the contract enabled gain and reference adaptation, with the tracking objective favoring higher tangential stiffness and slower reference advancement. In board writing, feedforward adaptation was enabled to regulate board-normal contact force. The problem is small and is solved once per adaptation cycle (100\,Hz on our robot), in closed form where the solution is analytic and by a bounded numerical solve otherwise. For a single constrained gain-only direction, the update selects the nearest feasible stiffness, accounting for gain--damping coupling (Supplementary Sec.~\ref{sec:supp-gain}). If the force interval is unattainable, clipping minimizes the scalar predicted violation within the gain bounds.

\begin{figure*}[!t]
\centering
\setlength{\abovecaptionskip}{2pt}
\includegraphics[width=\textwidth]{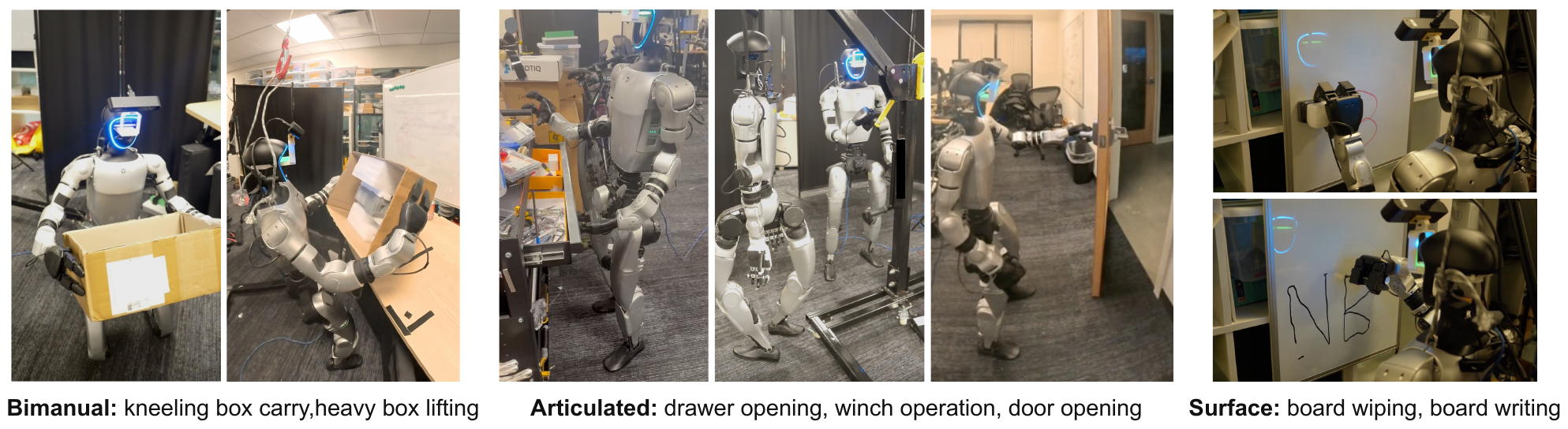}\unskip\par
\caption{Task coverage: bimanual transport, articulated-object manipulation and surface interaction.}
\label{fig:diverse-tasks}
\end{figure*}

\subsection{Realization on a Humanoid Robot}\label{sec:controller}

\textbf{Tunable task-space impedance control.}
The G1's quasi-direct-drive (QDD) arm actuators use low-ratio transmissions. Low reduction limits reflected motor inertia and facilitates backdriving~\cite{yu2020qdd}, making a compliant contact response easier to realize through joint torques. We therefore use classical Cartesian impedance control~\cite{hogan1985} to realize KPI's tunable response at each palm:
\begin{equation}\label{eq:imp}
\boldsymbol{\tau}=\mathbf{J}^{\top}\big(\mathbf{K}_p\mathbf{e}-\mathbf{K}_d\mathbf{v}+\mathbf{F}_{\mathrm{ff}}\big)
+\boldsymbol{\tau}_g+\boldsymbol{\tau}_{\mathrm{null}}.
\end{equation}
Here $\mathbf{J}$ is the palm Jacobian, $\mathbf{e}\in\mathbb{R}^6$ is the pose error relative to the reference conditioned by $\boldsymbol{\rho}$, and $\mathbf{v}=\mathbf{J}\dot{\mathbf{q}}$ is the arm-induced palm velocity for arm joints $\mathbf{q}$. The terms $\boldsymbol{\tau}_g$ and $\boldsymbol{\tau}_{\mathrm{null}}$ compensate gravity using measured base attitude and regulate redundant posture, respectively. The law runs at 500\,Hz on a wrench estimate refreshed at 200\,Hz. Translational stiffness defaults to 300\,N/m, is bounded to $[30,1000]$\,N/m, and takes 100\,N/m as the compliant target $\mathbf{K}^{\mathrm{soft}}_p$; damping follows the stiffness at critical damping, so the gain channel moves $\mathbf{K}_d$ with $\mathbf{K}_p$.

\textbf{Damping decomposition across control interfaces.}
On the G1, the motor drivers close joint-local PD feedback at a higher rate than the host communication loop. Implementing the full impedance law through the host torque-command channel produced oscillations in hardware tests: delayed velocity feedback acts on past motion, introducing phase lag that erodes dissipativity and stability margins~\cite{colgate1997}. The damping term in~\eqref{eq:imp} is $-\mathbf{J}^{\top}\mathbf{K}_d\mathbf{v}=-\mathbf{D}\dot{\mathbf{q}}$, and to exploit the faster local loop we decompose its joint-space matrix:
\begin{equation}\label{eq:damping-split}
\mathbf{D}=\mathbf{J}^{\top}\mathbf{K}_d\mathbf{J}
=\mathbf{D}_{\mathrm{diag}}+\mathbf{D}_{\mathrm{off}}.
\end{equation}
Each diagonal term depends only on the corresponding joint velocity and is implemented through the driver's damping gain $D_{ii}$. The host updates this gain as $\mathbf{J}$ and $\mathbf{K}_d$ change, while the driver uses fresh local velocity measurements between updates. Cross-joint terms remain host-computed as $-\mathbf{D}_{\mathrm{off}}\dot{\mathbf{q}}$, alongside the remaining impedance torques. With zero driver position gain and velocity reference, the combined paths recover Cartesian damping in the zero-delay limit. The split preserves task-space coupling while removing the host round-trip delay from joint-local damping; residual delay still constrains the achievable closed-loop response.

\textbf{Sensorless wrench estimation.}
Following Minimalist Compliance Control~\cite{mcc}, we correct actuator torque with a direction-dependent transmission-efficiency model and subtract gravity. Under the quasi-static approximation, regularized least squares through $\mathbf{J}^{\top}$ maps the resulting external-torque residual to a palm wrench. Friction effects cannot yet be fully eliminated, but with this correction the mean force-estimation error is expected to be below 5\,N, which is sufficient for the interaction tasks considered in this work.

\textbf{Whole-body control on a humanoid robot.}
SONIC~\cite{sonic} provides a three-point interface for head and wrist targets and an SMPL-based full-body interface. We use the former for agentic execution and the latter for teleoperation. In both modes, SONIC commands the legs and waist, KPI commands the arms, and the policy observes the actual joint state.
For agentic execution, SONIC's three-point targets and the impedance controller's hand references derive from the same commanded motion. For teleoperation, the SMPL motion goes directly to SONIC, while a parallel path uses General Motion Retargeting (GMR)~\cite{gmr} to map the same motion to robot joint positions. Forward kinematics then produces end-effector references for the Cartesian impedance controller. The contract itself is stated by the operator under teleoperation and written by the agent in agentic execution (Sec.~\ref{sec:agent}).

\subsection{Agentic Execution}\label{sec:agent}
A VLM identifies actionable geometry and composes motion from a language goal. With only egocentric RGB-D and proprioception, however, the contact conditions it plans against remain uncertain. Our framework pairs that reasoning with KPI: the agent supplies the nominal reference and the contract, and KPI adjusts the controller as the interaction unfolds (Fig.~\ref{fig:agent-pipeline}).

\textbf{Generating the reference and contract.}
Given an instruction and current observations, the \emph{Analyzer} requests relevant geometry from the \emph{Detector}, whose VLM localizes task-relevant features in RGB images. The corresponding depth measurements map these locations into 3D, giving grasp points and geometric parameters such as trajectory radii. A \emph{motion core} is a reusable, parameterized motion generator, such as linear translation or arc motion. Analyzer organizes the task into \emph{stages}. Each stage names a motion core and its parameters, which define the nominal reference, together with a contract $\sigma$ (Sec.~\ref{sec:layer}). The \emph{Verifier} checks the proposal against observations and robot constraints, with kinematic and collision checks; Analyzer and Verifier are role-conditioned calls to the same VLM. Across tasks the agent composes shared motion cores and contracts without task-specific code.

\textbf{Physical adaptation and task-level feedback.}
While the \emph{Executor} runs a stage, KPI obtains the selected parameters through~\eqref{eq:adaptation} from the interaction state, so the objectives and force constraints guide the response to changing geometry and loading without the VLM specifying each parameter update.

After each stage, Verifier evaluates fresh observations against the intended outcome, including evidence that the execution met difficulty, such as persistent tracking error. Analyzer then revises the remaining reference and contract from the current state, retaining completed stages, and the revision is verified before execution resumes; Fig.~\ref{fig:agent-pipeline}(b) follows one such revision. Completion is judged against the environmental goal, not the end of a trajectory.

\begingroup
\setlength{\intextsep}{3pt plus 1pt minus 1pt}
\setlength{\textfloatsep}{3pt plus 1pt minus 1pt}
\clubpenalty=10000
\widowpenalty=10000
\raggedbottom

\section{Experiments}\label{sec:experiments}

KPI supports three task categories (Fig.~\ref{fig:diverse-tasks}): bimanual transport (box placement and carrying during kneeling, standing, and running); articulated-object manipulation (drawer and door opening, and winch operation); and surface interaction (board wiping and writing). We evaluate zero-shot agentic execution on winch operation, door passage, and box transport and placement.

\begin{figure}[!htbp]
\centering
\setlength{\abovecaptionskip}{2pt}
\includegraphics[width=\columnwidth]{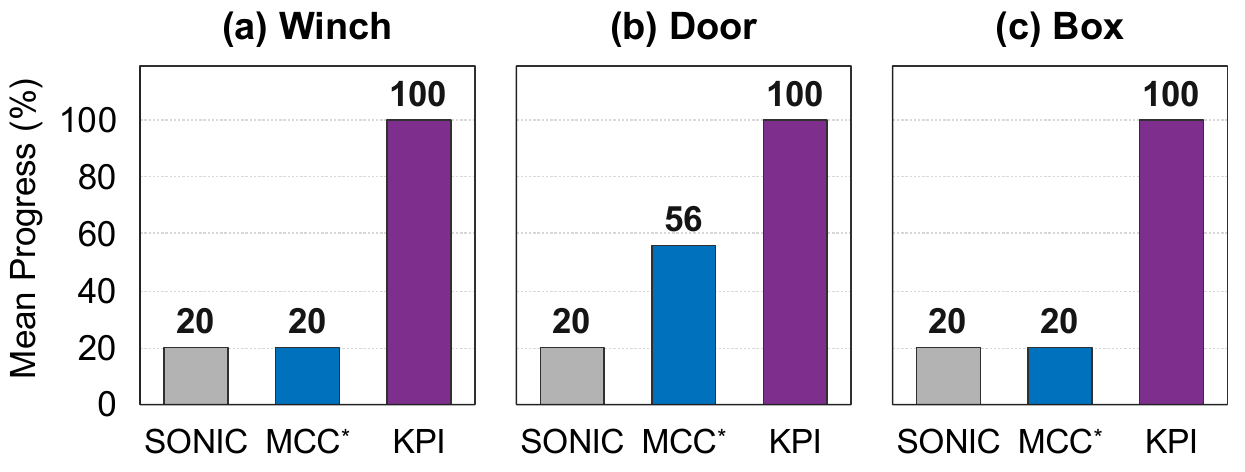}\unskip\par
\caption{Mean progress for agentic executions.}
\label{fig:agent-progress}
\end{figure}

Our analyses examine (1) whether KPI improves zero-shot agentic execution over tracking and compliant-control baselines, (2) its effects on physical interaction and the contribution of online adaptation, and (3) whether these benefits extend across tasks and whole-body motions.

We use a Unitree~G1 with Dex3 hands, egocentric RGB-D, and proprioception, but no wrist force sensors. All conditions share SONIC's whole-body tracker~\cite{sonic} (three-point tracking for agentic execution; SMPL tracking for teleoperation), with arm-controller interfaces as in Sec.~\ref{sec:controller}. GPT-6 Astra implements Analyzer, Detector, and Verifier.

We compare four conditions: \emph{SONIC}, SONIC~\cite{sonic}'s native joint PD control; \emph{MCC$^{*}$}, MCC's sensorless admittance controller~\cite{mcc} integrated with SONIC through the whole-body control scheme in Sec.~\ref{sec:controller}; \emph{KPI-fixed}, our impedance realization without controller-parameter adaptation; and \emph{KPI}, the full kernel with online adaptation.

Each zero-shot agentic trial starts from one instruction, without task-specific code or a manually supplied contract. SONIC, MCC$^{*}$, and KPI share the agentic system and prompts, except that MCC$^{*}$ requests controller stiffnesses rather than contract parameters. SONIC uses KPI's agent interface but ignores the generated contract during execution. For force-requiring contacts, we offset the nominal position reference 10\,cm from the target contact point along the intended force direction by default.

\subsection{Winch Operation}
\label{sec:exp-winch}
\label{sec:exp-ablation}

\textbf{Task and challenges.}
The robot must grasp and rotate a winch handle to hoist another G1 robot (Fig.~\ref{fig:traces}). The handle moves along a constrained one-DoF circular path. Resistance varies within each revolution and increases as the other robot is lifted, requiring continual adjustment of tangential force. Excessive force in constrained directions can impede rotation, push the operating robot backward, or cause loss of grasp.

\textbf{Agentic execution.}
We request four clockwise revolutions at a moderate rate. Analyzer selects circular motion, assigning \emph{tracking} along the path tangent and \emph{constrained} requirements to the other translational directions. It also assigns a \emph{compliant} requirement to hand rotation, realized with low rotational stiffness. Detector identifies a grasp point and estimates trajectory geometry from the pivot pin and handle position.

We define five evaluation stages: grasping the detected point, then one stage per completed revolution; progress is the fraction completed. Across five trials per method, KPI completes all four revolutions, achieving 100\% mean progress (Fig.~\ref{fig:agent-progress}). SONIC and MCC$^{*}$ grasp the handle but do not exceed half a revolution, yielding 20\% mean progress.

\begin{figure}[!htbp]
\centering
\setlength{\abovecaptionskip}{2pt}
\input{figures/experiment_winch}\unskip\par
\caption{Winch operation. (a) Operation sequence. (b) KPI and KPI-fixed's hand trajectories. (c) MCC$^{*}$ failure case.}
\label{fig:traces}
\end{figure}

\textbf{Analysis.}
Two coupled effects enable sustained rotation. (1) When high resistance reduces hand speed well below the desired speed (Fig.~\ref{fig:traces}(b), left), the optimization slows the conditioned reference and increases tangential stiffness to maintain tracking. Limiting reference lead also reduces misalignment between tracking force and the locally admissible tangent. (2) Force caps in constrained directions limit the predicted response to avoid excessive loading against the mechanism.

KPI-fixed holds all VLM-selected controller parameters fixed. At a high-resistance point (Fig.~\ref{fig:traces}(b), right), the hand stalls while the reference advances; the tracking force becomes increasingly misaligned with the path tangent until the hand reverses.

Across baseline trials, excessive contact forces commonly cause unintended body motion or loss of grasp. In the MCC$^{*}$ failure case (Fig.~\ref{fig:traces}(c)), the hand stalls in the lower circle while continued reference advance generates an upward tracking force that deflects the wrist.

\subsection{Door Opening and Passage}
\label{sec:exp-door}

\textbf{Task and challenges.}
The robot must grasp the handle, open the door while stepping backward to avoid its swing, then open it further and walk through (Fig.~\ref{fig:agentloop}). The main challenge is incomplete visibility: although the VLM can infer the required motions, the robot cannot initially observe the full door and therefore cannot accurately estimate its opening trajectory.

\textbf{Agentic execution.}
We instruct the robot to pull the door toward itself and take two steps through the doorway. The agent selects a handle grasp point and assigns rotational \emph{tracking} to turn the handle and release the latch. With only part of the door visible, it combines backward stepping with a leftward hand reach and assigns \emph{compliant} requirements in all directions for opening. For evaluation, we define five progress stages (Fig.~\ref{fig:agentloop}): grasping, unlatching, stepping back, swinging the door open, and walking through.

Completion requires both feet and the torso to pass through the doorway. Over five trials per method (Fig.~\ref{fig:agent-progress}), KPI completes all five, SONIC fails to unlatch in every trial, and MCC$^{*}$ completes the passage once, with substantial variation across the remaining trials.

\begin{figure}[!htbp]
\centering
\setlength{\abovecaptionskip}{2pt}
\input{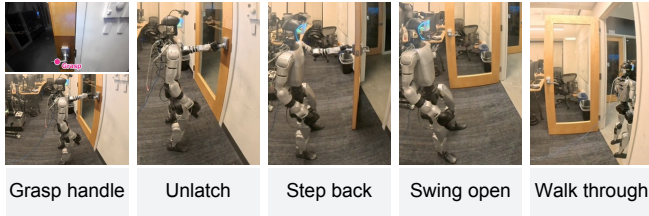}\unskip\par
\caption{Door opening and passage: grasp the handle, unlatch, step back, swing open, and walk through.}
\label{fig:agentloop}
\end{figure}

\textbf{Analysis.}
KPI's adaptive tracking reliably turns the handle to release the latch. MCC$^{*}$ also succeeds with the relatively high rotational stiffness selected by the VLM, but its admittance controller modifies the tracking reference using estimated forces. Estimation errors cause slight oscillations and reduced tracking accuracy, leading to a missed grasp in one trial.

After unlatching, the robot steps backward while pulling the handle. Because the commanded hand motion conflicts with the hinge constraint, compliance is essential to accommodate the mismatch. MCC$^{*}$ achieves compliance by adjusting the motion reference from estimated forces, but responds more slowly to external forces than KPI in our implementation. The hand can therefore repeatedly impact the handle, consistent with \emph{contact instability} and bouncing reported for admittance-controlled interaction with stiff environments~\cite{keemink2018admittance}. These oscillations make door pulling unreliable, and MCC$^{*}$ completes only one of five trials.

\subsection{Box Transport and Load Robustness}
\label{sec:exp-box}

\textbf{Task and challenges.}
The robot must transport a box between two tables while maintaining a bimanual grasp through posture changes. Each three-fingered Dex3 hand must maintain multiple sidewall contacts, yet small hand--box orientation changes can break individual finger contacts. The hands must maintain inward forces while accommodating relative-pose changes during turns and squats. These forces must prevent slip without crushing the box or overheating the robot.

We test three box masses: Light (0.85\,kg), Medium (1.85\,kg), and Heavy (3\,kg). To limit overheating risk during prolonged agentic execution, the full sequence uses only Light; separate single-step lifting tests assess load robustness across all three masses.

\textbf{Agentic execution.}
The destination table is initially out of view behind the starting table, and the carried box obscures the view below. We therefore prompt the robot to pick up the box, turn $180^\circ$, squat 10\,cm, and place it on the table. The agent still selects motion cores, specifies the interaction contract, chooses grasp points, and plans hand trajectories to avoid unintended collisions with table edges and the box.

Progress is evaluated over five stages (Fig.~\ref{fig:box-robustness}(a)). Across five trials per method, KPI achieves 100\% mean progress; SONIC and MCC$^{*}$ each achieve 20\%, failing to lift the box (Fig.~\ref{fig:agent-progress}).

\begin{figure}[!htbp]
\centering
\setlength{\abovecaptionskip}{2pt}
\input{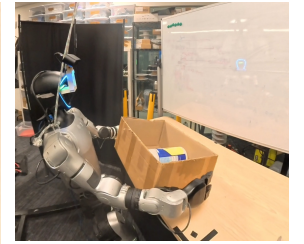}\unskip\par
\caption{Box transport and load robustness. (a) The five stages of transport and placement. (b) Maximum liftable load category. (c) SONIC fails to maintain multi-point contact with the box walls, preventing lifting.}
\label{fig:box-robustness}
\end{figure}

\textbf{Analysis.}
To assess load adaptation, stationary lifting tests place the load at far, middle, and near positions inside the box (Fig.~\ref{fig:box-robustness}(b)). The robot lifts for 3\,s, then holds for 3\,s; success requires the box to remain fully clear of the table throughout the hold. Liftable mass depends on hand--box friction; our ordinary shipping carton has packing tape on its sides and no antislip treatment, yielding relatively low-friction contacts.

In both evaluations, SONIC and MCC$^{*}$ cannot lift the box because they fail to maintain stable multi-point contact between both hands and the sidewalls (Fig.~\ref{fig:box-robustness}(c)). SONIC's wrist-located tracking keypoints during training may leave its policy insufficiently resistant to hand deflection under contact forces. MCC$^{*}$ uses admittance control, where end-effector torque-estimation errors can induce unintended hand rotation under load.

KPI-fixed also performs poorly with Medium and Heavy loads: excessive force output triggers motor errors in its failed trials. Despite being informed of the 10\,cm reference offset, the agent selects stiffnesses of 500 and 600\,N/m, respectively. For KPI, it instead selects force ranges of 25--40\,N (Medium) and 35--50\,N (Heavy). With closed-loop adaptation, these specifications yield lower contact forces than KPI-fixed, suggesting that the VLM selects force ranges more appropriately than stiffness values for this task.

\subsection{Board Writing}\label{sec:exp-writing}

\textbf{Task and challenges.}
Writing requires sustained contact force between the pen and the whiteboard. Insufficient force causes contact loss, whereas excessive force can impede the pen's motion.

\textbf{Experiment and analysis.}
We compare all four methods from the same initial hand pose relative to the whiteboard. KPI-fixed and MCC$^{*}$ use the same feedforward force. KPI treats the feedforward force as an optimization variable, with its contract's force target set to the same value. KPI and KPI-fixed use zero stiffness along the board normal; the remaining stiffness settings are matched across KPI, KPI-fixed and MCC$^{*}$.

Fig.~\ref{fig:writing-results} compares typical writing results with the nominal \emph{NB} reference. KPI and KPI-fixed both produce recognizable \emph{N} and \emph{B} shapes. SONIC's lower tracking accuracy may stem from a relatively soft response under contact. MCC$^{*}$'s poorer result may reflect instability in its admittance response driven by estimated forces. KPI's mean estimated contact force is closer to the target than KPI-fixed's, but this improvement has little visible effect on the written pattern.

\begin{figure}[!htbp]
\centering
\setlength{\abovecaptionskip}{2pt}
\input{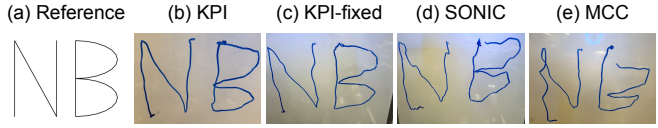}\unskip\par
\caption{Board-writing results using different methods.}
\label{fig:writing-results}
\end{figure}

\subsection{Additional Tasks and Whole-Body Motions}\label{sec:exp-generality}\label{sec:exp-teleop}

Through teleoperation, KPI also supports drawer opening and box carrying while running, kneeling and standing up (Fig.~\ref{fig:diverse-tasks}). A whole-body tracker with greater waist and leg mobility and robustness to external forces could extend our system to even more challenging tasks.

\endgroup

\FloatBarrier
\section{Conclusion}\label{sec:conclusion}
KPI provides an explicit, promptable kernel between the commander and an unmodified whole-body tracker, adapting the arm controller online at contact rate according to the commander’s interaction contract. From one instruction and with no task-specific code, a vision-language agent turned a hand-winch until a second robot was lifted off the ground, opened a door and walked through it, and carried a box between two tables; a teleoperator drove the same interface to open a drawer and to carry a box while kneeling, standing and running.

The kernel does not choose the objectives. They come from a person or a model, and a jam that no contract anticipates appears only as a stalled stage. The wrench estimate is quasi-static, and its error floor rises with stiffness and gait; joint torque sensors or tactile palms would lower it and admit lighter contact. By design, any task-space arm can run the kernel and any trajectory source can call it.

\bibliographystyle{IEEEtran}
\bibliography{refs}

\clearpage
\onecolumn
\setcounter{section}{0}
\setcounter{subsection}{0}
\setcounter{equation}{0}
\setcounter{figure}{0}
\setcounter{table}{0}
\renewcommand{\thesection}{S\arabic{section}}
\renewcommand{\theequation}{S\arabic{equation}}
\renewcommand{\thefigure}{S\arabic{figure}}
\renewcommand{\thetable}{S\arabic{table}}
\renewcommand{\sfdefault}{lmss}
\begin{center}
  {\Large\bfseries Supplementary Material}
\end{center}
\providecommand{\KPIImpedanceEquation}{\eqref{eq:imp}}
\providecommand{\KPIObjectiveEquation}{\eqref{eq:contract-objective}}
\section{Local Model and Parameter Update}\label{sec:supp-gain}

All predictions are local to the current state and expressed in a consistent contact frame. We distinguish the immediate effect of a parameter update from the short-horizon motion response. Both follow from the impedance relation in Eq.~\KPIImpedanceEquation, without identifying environmental stiffness. The approximation assumes slowly varying contact over the prediction interval; impacts and strongly inertial interaction are outside its scope.

Along direction $i$, let $r_i(\boldsymbol{\rho})$ be the conditioned reference, $x_i$ and $v_i$ the measured end-effector coordinate and velocity, $k_i$ and $b_i$ the controller stiffness and damping, and $u_i$ the feedforward component. Set $e_{i,t}=r_i(\boldsymbol{\rho}_t)-x_{i,t}$. Signed loads use the robot-on-environment convention, with external-wrench estimates converted accordingly. All coefficients below are evaluated at the current state.

For the immediate force prediction, we hold measured pose and velocity fixed. The first-order parameter-induced load increment is
\begin{equation}\label{eq:supp-load-model}
\begin{aligned}
\delta w_i
&=e_{i,t}\Delta k_i+k_{i,t}\Delta r_i-v_{i,t}\Delta b_i+\Delta u_i,\\
\tilde f_i&=\hat f_i+\delta w_i.
\end{aligned}
\end{equation}
Here $\Delta r_i=(\partial r_i/\partial\boldsymbol{\rho})_t\Delta\boldsymbol{\rho}$, and the prescribed gain--damping relation is differentiated when forming the local update. Unselected channels have zero increments. The measured wrench anchors the prediction; it is not an optimization variable.

For the tracking objective, let $z_i(s)$ denote the incremental motion caused by the same parameter perturbation. We fix the prediction horizon at $h=10$\,ms, one adaptation period. Holding the current external load fixed over this horizon, a damping-dominated approximation gives
\[
b_{i,t}\dot z_i+k_{i,t}z_i=\delta w_i,\qquad z_i(0)=0.
\]
For a tracking direction with $k_{i,t},b_{i,t}>0$,
\begin{equation}\label{eq:supp-tracking-model}
\begin{aligned}
g_i(h)&=\frac{1-\exp(-k_{i,t}h/b_{i,t})}{k_{i,t}},\\
\tilde x_{i,t+h}&=x_{i,t}+h v_{i,t}+g_i(h)\delta w_i.
\end{aligned}
\end{equation}
Thus $g_i$ depends only on the robot's controller parameters and the prediction horizon, not on an estimated environmental compliance. The two errors in Eq.~\KPIObjectiveEquation{} are
\begin{equation}\label{eq:supp-tracking-errors}
\tilde{\mathbf e}_i=
\begin{bmatrix}
x^\star_{i,t+h}-\tilde x_{i,t+h}\\
r_{i,t+h}(\boldsymbol{\rho})-\tilde x_{i,t+h}
\end{bmatrix}.
\end{equation}
Equation~\eqref{eq:supp-load-model} predicts an immediate load change, whereas Eq.~\eqref{eq:supp-tracking-model} supplies a short-horizon tracking surrogate; they are not simultaneous exact predictions of a future contact state. Fresh measurements replace both predictions at the next update. The corresponding matrix equations retain controller coupling, including both arms. Subspace bounds use the norm of the projected vector force prediction; rotations use local angular coordinates.

\medskip
For the \textbf{single-direction gain-only special case}, reference and feedforward are fixed. Writing $\Delta b_i=\beta_i\Delta k_i$ in the local linearization gives
\[
\begin{aligned}
\tilde f_i(k)&=\hat f_i+a_i(k-k_{i,t}),\\
a_i&=e_{i,t}-v_{i,t}\beta_i.
\end{aligned}
\]
A constrained specification adds no tracking or stiffness-preference objective. In this scalar local reduction, the positive-definite regularizer is a positive multiple of $(k-k_{i,t})^2$, so the feasible solution is the stiffness closest to its current value. For $a_i\ne0$,
\begin{equation}\label{eq:gain}
\begin{aligned}
f_{i,b}&=\operatorname{clip}\!\left(\hat f_i,f_i^{\min},f_i^{\max}\right),\\
k_{i,t+1}&=\operatorname{clip}\!\left(k_{i,t}+\frac{f_{i,b}-\hat f_i}{a_i},\ \underline{k}_i,\ \bar{k}_i\right).
\end{aligned}
\end{equation}
This retains stiffness inside the force band and reaches the violated boundary when attainable. Otherwise, clipping gives the parameter endpoint with the smallest predicted violation: a bounded fallback, not satisfaction of the original hard constraint. If $a_i=0$, gain has no local force authority and stiffness is retained.

\FloatBarrier

\end{document}